\documentclass[10pt,twocolumn,letterpaper]{article}
\usepackage[pagenumbers]{cvpr}

\usepackage{amsmath,amsfonts,bm}

\def\eqref#1{equation~\ref{#1}}

\def\1{\bm{1}}

\newcommand{\test}{\mathcal{D_{\mathrm{test}}}}

\DeclareMathAlphabet{\mathsfit}{\encodingdefault}{\sfdefault}{m}{sl}
\SetMathAlphabet{\mathsfit}{bold}{\encodingdefault}{\sfdefault}{bx}{n}

\definecolor{cvprblue}{rgb}{0.21,0.49,0.74}
\usepackage[pagebackref,breaklinks,colorlinks,allcolors=cvprblue]{hyperref}

\usepackage{hyperref}
\usepackage{graphicx}
\usepackage{caption}
\usepackage{booktabs} 
\usepackage{multirow}
\usepackage{pifont}%
\newcommand{\cmark}{\ding{51}}%
\newcommand{\xmark}{\ding{55}}%
\usepackage{sidecap}

\usepackage[linesnumbered,ruled,vlined]{algorithm2e}
\usepackage{xcolor}

\def\paperID{2239} %
\def\confName{CVPR}
\def\confYear{2025}

\definecolor{mydarkblue}{RGB}{0,0,160} %

\SetCommentSty{mycommfont}
\SetKwInOut{Parameter}{Hyperparams}

\title{DiSciPLE: Learning Interpretable Programs for Scientific Visual Discovery}

\author{
Utkarsh Mall$^{1}$ \hspace{0.05cm} Cheng Perng Phoo$^{2}$  \hspace{0.05cm} Mia Chiquier$^{1}$ \hspace{0.05cm}
Bharath Hariharan$^{2}$ \hspace{0.05cm}  Kavita Bala$^{2}$ \hspace{0.05cm} Carl Vondrick$^{1}$\\
         $^{1}$Columbia University \qquad
        $^{2}$ Cornell University\\
        \small{\enspace Correspondence: \tt{um2171@columbia.edu}
}\\
\href{https://disciple.cs.columbia.edu/}{disciple.cs.columbia.edu}
}

\newcommand{\best}[1]{\textbf{\textcolor{red}{#1}}}
\newcommand{\sota}[1]{\emph{\textcolor{blue}{#1}}}

\def\disciple{DiSciPLE}

\begin{document}

\maketitle
\begin{abstract}

Visual data is used in numerous different scientific workflows ranging from remote sensing to ecology. As the amount of observation data increases, the challenge is not just to make accurate predictions but also to understand the underlying mechanisms for those predictions. 
Good interpretation is important in scientific workflows, as it allows for better decision-making by providing insights into the data. 
This paper introduces an automatic way of obtaining such interpretable-by-design models, by learning programs that interleave neural networks. We propose \disciple~(Discovering Scientific Programs using LLMs and Evolution) an evolutionary algorithm that leverages common sense and prior knowledge of large language models (LLMs) to create Python programs explaining visual data. Additionally, we propose two improvements: a program critic and a program simplifier to improve our method further to synthesize good programs. On three different real-world problems, \disciple~learns state-of-the-art programs on novel tasks with no prior literature. For example, we can learn programs with 35\% lower error than the closest non-interpretable baseline for population density estimation. The supplementary material can be found at: \href{https://disciple.cs.columbia.edu/pdf/supplementary.pdf}{https://disciple.cs.columbia.edu/pdf/supplementary.pdf}

\end{abstract}

\section{Introduction}
\label{sec:intro}

Many modern scientific workflows are built on top of visual data. 
Researchers in remote sensing, climate science, ecology, and other sciences use images as a window into our world to estimate population density, the amount of biomass, poverty indicators, and so on. There is a massively increasing volume of visual data, be it from an ever-expanding set of satellites, widespread camera traps, or images uploaded on the web, that is available to domain experts, and computer vision has the potential to meaningfully assist scientists in using it for scientific insight.

Scientific applications of computer vision, however, are demanding tasks because we want models that not only predict outcomes but also reveal underlying mechanisms. For example, a researcher who studies demography may be able to train excellent predictive models that learn the relationships between a satellite image and the population.
However, understanding why certain regions are densely populated is crucial for urban planning and policy decisions --- black-box predictions offer no interpretation of what makes a region have a high population. 
Scientists themselves want to derive insight from the models, not just predict.

\begin{figure}[t!]
    \centering
\includegraphics[width=\linewidth]{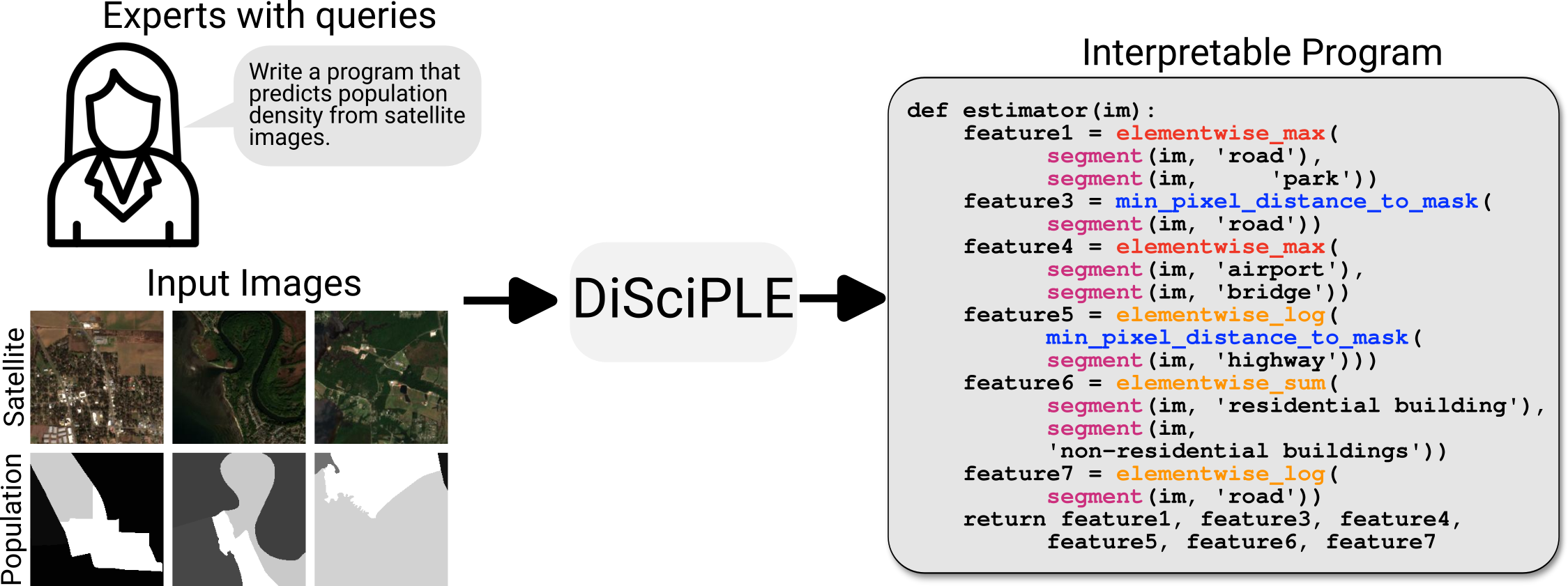} %
\caption{We introduce a framework to discover interpretable, predictive programs for scientific computer vision tasks.}
\label{fig:teaser}
\end{figure}

While there have been many works that train interpretable vision models (for example concept-bottlenecks~\citep{koh20concept,menon2022visual}), these models are often limited to simple functions of primitive concepts, such as bag of words.
These simple functions do not scale to the realistic complexity of our visual world and the complex relationships between its many rich scientific indicators, resulting in poor accuracy. 
A promising direction to model rich relationships without foregoing interpretability is to learn \emph{programs} on top of conceptual primitives.
Recently, code generation methods such as ViperGPT and VisProg \cite{suris-23,gupta2023visual} have demonstrated that large language models are able to synthesize programs with competitive performance on many vision tasks, and the representations are interpretable by construction because programs are human-readable. However, while these methods work well for established vision tasks, they often fail to generalize to scientific applications of computer vision because the tasks are new and outside the scope of the training data on the internet. LLMs lack the requisite knowledge to answer novel or domain-specific questions, and as we will show, directly applying such zero-shot code generation methods on scientific domains is not effective.

How can we automatically discover accurate and interpretable programs from the volumes of visual data in scientific applications? This paper introduces DiSciPLE, a framework for \textbf{Di}scovering \textbf{Sci}entific \textbf{P}rograms  using \textbf{L}LMs
and \textbf{E}volution. Given a large dataset of images, our approach learns to synthesize a program for solving the task. As the name suggests,~\disciple~introduces an evolutionary search algorithm that starts with zero-shot programs from LLMs and iteratively improves them over the dataset. The discovered programs are able to interleave neural networks, in particular open-world segmentation foundation models, for segmentation with logical and mathematical operations, enabling powerful predictions while also being interpretable. Our framework makes several improvements to integrate evolutionary search with LLMs, using program simplification and critics to provide fine-grained guidance that accelerates program search.

In three different scientific applications of computer vision, ~\disciple~is able to learn state-of-the-art programs for novel tasks that have no prior documented solutions in existing literature. Our approach significantly outperforms neural networks at estimating population density from satellite imagery. Our method also obtains strong out-of-distribution performance at estimating a region’s biomass,  generalizing to geographical regions outside of the training set significantly better than all other baselines. 

Our contributions are:
\begin{itemize}
    \item We introduce a novel framework \textbf{\disciple}, that can produce interpretable, reliable, and sample-efficient programs for scientific discovery.
    \item We present two key components: a critic and a program simplification method to \disciple~that can further improve the search resulting in better programs.
    \item We propose benchmarks for the task of scientific visual discovery containing real-world high-dimensional visual data for three problems in two different domains. We also apply \disciple~on these benchmarks and show that our learned programs are more interpretable, reliable, and data-efficient compared to baselines.    
\end{itemize}

\section{Related Works}
\label{sec:related_works}
\paragraph{Concept bottlenecks.}
Concept bottleneck~\citep{koh20concept,yang2023language,oikarinen2023label} is an approach used to create interpretable-yet-powerful classifiers. 
The key idea is to train a deep model to predict a set of low-level concepts or bottlenecks and then learn a linear classifier. 
Such concept bottlenecks have the basis of methods in several areas such as fine-grained recognition \citep{ferrari2007learning, huang2016part,zhou2018interpretable,tang2020revisiting} and zero-shot learning ~\citep{lampert-13,akata-15,kodirov-17}.
However in order to train these models, expensive data is needed to be collected for the bottleneck concepts themselves. 
One way to reduce this annotation cost is to sequentially ask questions in an information-theoretically optimized way ~\citep{chattopadhyay2024bootstrapping, chattopadhyay2024information}.
Researchers have also automated this pipeline by using large-language models as a knowledge base to propose concept bottleneck models~\citep{menon2022visual,pratt2023does, han2023llms}. 
\cite{llmmutate-24} proposed an evolutionary algorithm with LLMs as the mutation operation to discover interpretable concept bottleneck models without prior information. While these models are interpretable, they are very simple in terms of expressive power.  
In this work, we instead evolve more expressive programs than a bag of words, while being interpretable.

\paragraph{Symbolic regression.}
Symbolic regression (SR)~\citep{cranmer2023interpretable} is a technique for learning equations through evolutionary search.
Several methods have been proposed to improve the search efficiency~\citep{makke2024interpretable}, however, most SR techniques cannot solve problems beyond simple mathematical formulas, with simple mathematical primitives.
This is partly because the search space of solutions is combinatorially too large. 
As a result, SR methods fail to work for images, which are too high-dimensional.
Our method instead focuses on problems with high-dimensional visual data, by leveraging visual foundation models as primitives.
This results in models that are better performing while being interpretable.
Like our approach, recent work on SR~\citep{grayeli2024symbolic, shojaee2024llm,merler2024context,li2024automated} has also looked at using LLMs to better guide the search.
However, these methods are only tested on lower-dimension mathematical problems for formula discovery, with a limited set of primitives. 
We instead propose an approach that is complementary to these methods.
Methods for SR cannot be applied directly in higher-dimensional open-world visual problems, on the other hand on low-dimensional problems existing tools for SR~\citep{grayeli2024symbolic} would perform better than~\disciple.
The focus of this work is on such real-world problems, where the primitive functions are more complex than mathematical operations and can even be open-world, for example, a text-to-image segmentation. 

\paragraph{Neuro-Symbolic Program Learning~\citep{mao2019neuro,dongneural}}is another avenue for learning programs for observation datasets or question answering. 
These methods typically try to learn both discrete program structures together with neural networks. 
However, since this optimization is non-differentiable these methods require reinforcement learning~\citep{johnson2017inferring} or complex non-differentiable optimization techniques ~\citep{ellis2021dreamcoder}.
The hard optimization issue makes the problem of learning programs sample inefficient in real-world settings.
We alternatively use LLMs ability to program to better guide the search for such programs.

\paragraph{Program synthesis with LLMs.}
Several works have utilized LLM coding ability in different applications such as VQA~\citep{suris-23, gupta2022visual} and robot manipulation~\citep{liang2023code}. 
While the zero-shot inferred code work very well on domains well-known to the internet, they tend to perform poorly on problems in scientific domains, as shown by our results. 

\paragraph{Scientific applications.}
Researchers in numerous scientific domains have used machine learning tools to build predictive models for their quantities of interest. 
In this work, we focus on two such scientific domain of: demography and climate science.
For both these domains, we use remote sensing vision language foundation models as powerful primitives, along with mathematical, logical and image operators.
In demography, we focus on the problems of socioeconomic indicator prediction~\citep{yong-24}, namely population density and poverty estimation~\citep{metzger-24,xie2017mapping}.
Similarly in climate science we focus on the problem of aboveground biomass prediction (AGB)~\citep{nathaniel2023above}.

\section{Methodology}
\label{sec:method}
Our key contribution is a program search framework that leverages LLMs to perform evolutionary search. In \cref{ssec:formulation} we formalize the problem of program search.
In \cref{ssec:evolution} we present our method of incorporating LLMs in the evolutionary search framework. 
Finally, in \cref{ssec:setpred}, \ref{ssec:critic} and \ref{ssec:simplification}, we discuss the improvements to this framework to speed-up the search.

\begin{figure*}[ht]
    \centering
\includegraphics[width=\linewidth]{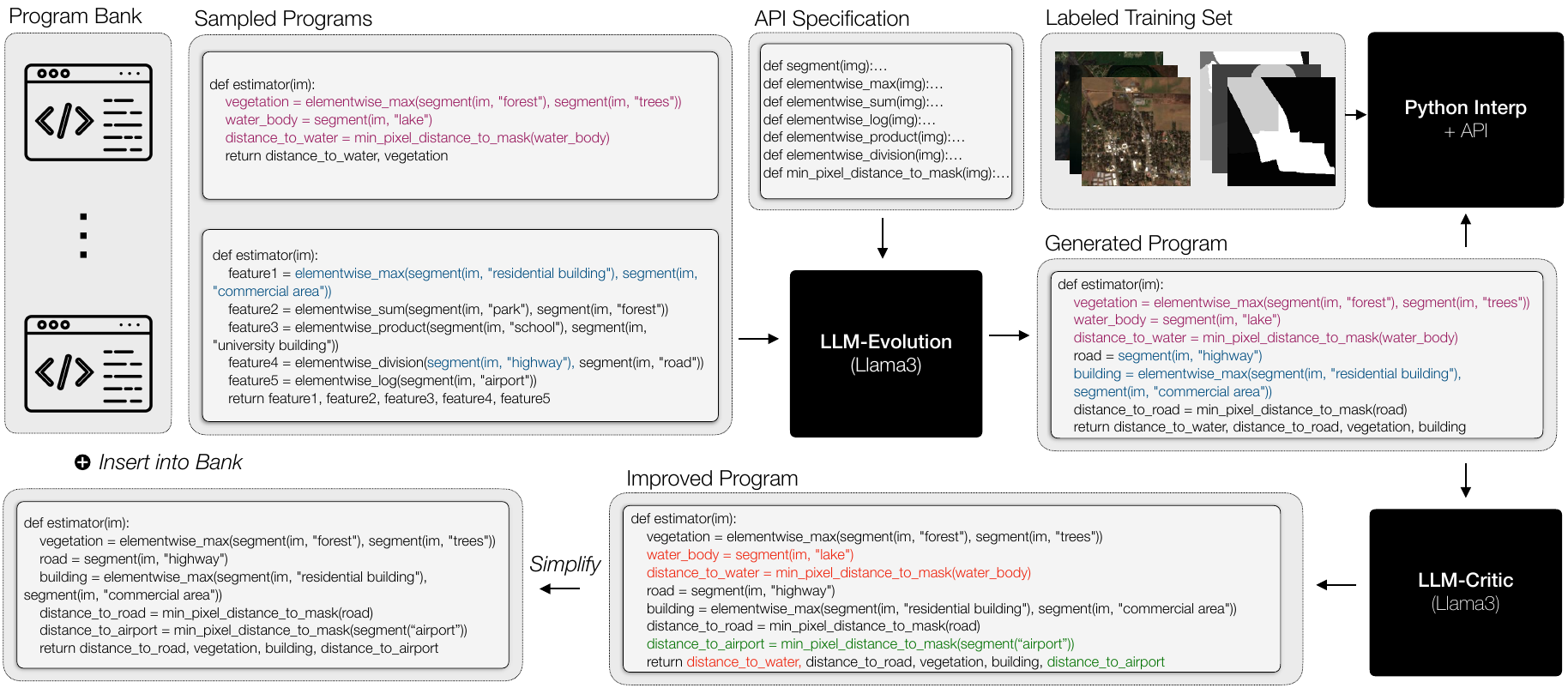}
\caption{Overview of our evolutionary algorithm with \textit{critic} and \textit{simplification}. We start with an initialized bank of program trying to solve a task. From this bank we sample pairs of programs based on their fitness score and perform crossover/mutations over them to produce new programs. The generated program is further improved by passing it through a critic and then an analytical simplification step. This program is then evaluated and put in the next generation of program bank. The evaluation score of the program is used to determine the fitness for the next generation of evolution.}
\label{fig:pipeline}
\end{figure*}

\subsection{Problem Formulation}\label{ssec:formulation}
Our system receives as input a \textbf{dataset} $\mathcal{D} = \{(x_1, y_1), (x_2, y_2), \dots, (x_n, y_n)\}$ consisting of inputs $x_i \in X$ and quantities of interest $y_i \in Y$.
For example, when estimating geospatial indicators like poverty, $x_i$ may be a latitude and longitude, along with other metadata about that location.
The system must produce an \emph{interpretable program} $P: X \rightarrow Y$ that maps inputs to corresponding outputs.
As is typical with standard supervised learning, we also assume a loss function or \textbf{metric} $\mathcal{M}$ that measures how good a particular prediction is, and seek a program that yields the best evaluation score:
\begin{equation}
\label{eq:score}
s(P; \mathcal{D}) = \frac{1}{n}\sum_{i=1}^{n}\mathcal{M} (P(x_i), y_i)    
\end{equation}
For example, in the case of population density a good metric used by domain experts is L2 error over log \textit{i.e.} $\mathcal{M}(y', y) = ||log(y')-log(y)||_2$~\citep{metzger-22,metzger-24}.

For our programs to be interpretable, they must put together modules or \textbf{primitives} in an interpretable way.
We conceptualize these primitives as a library of functions $\mathcal{F} = \{f_1, f_2 \dots f_k\}$, that can be used to construct a program $P$.
Given that we are analyzing visual data, a key primitive will be an open-vocabulary recognition model that is applied to any imagery associated with a data point.
For example, when estimating poverty for a location, we can define a primitive that 
queries the satellite images available at that location and uses an open-vocabulary recognition engine such as GRAFT~\cite{mall2023graft} to detect/segment various concepts.
Recent advances in recognition have produced such foundation models for a range of modalities~\cite{mall2023graft, stevens2024bioclip}. 
In addition, we will assume basic mathematical, logical and image operators such as
logarithms, elementwise maximum, or a distance transform.
We note that the set of primitives is often not specific to the task and can be shared across a range of problems in a domain.
That said, the set of primitives can be expanded upon if needed by domain experts for particular problems; for example, a climate scientist may want to include a function that can look up the average temperature at a particular place and time.

Finally, to enable us to search the space of programs effectively and leverage the conceptual understanding of LLMs, we assume that we have a natural language name or description $descr$ of the quantity of interest $Y$. For example, this may be the phrase ``population density". 
As we will see below, this information will be useful in guiding the LLM to search the space well.

Putting everything together, our proposed system, \disciple~(\textbf{Di}scovering \textbf{Sci}entific \textbf{P}rograms  using \textbf{L}LMs
and \textbf{E}volution), takes as input the dataset $\mathcal{D}$, the metric $\mathcal{M}$, the set of primitives $\mathcal{F}$ and the textual description $descr$. It produces an interpretable and accurate program $P$ that maps inputs $X$ to the output quantity of interest $Y$.

We next describe our proposed system.

\begin{algorithm}[t!]
\caption{DiSciPLE's learning loop}
\label{alg:algorithm}

\DontPrintSemicolon %

\KwIn{Observation set $\mathcal{D} = \{(x_1, y_1), (x_2, y_2), \dots, (x_n, y_n)\}$, metric $\mathcal{M}$, an objective prompt $p_o$, a set of primitive function $\mathcal{F} = \{f_1, f_2 \dots f_k\}$} 
\Parameter{Mutation probability $\rho_m$, total number of generations $T$, population size $M$, crossover $p_c$ and and mutation $p_m$ prompts.}
\KwOut{A program $P^*$ in the form of a program that explains the observations best.} %

$B^0 \gets \{\}$ \tcp{Initialize a programs bank} 

\For{$i = 1,\ldots,M$}{
    $P^0_i \gets \mathcal{LLM}(p_o)$
   
    $B^0 \leftarrow B^0 \; \cup \; \{ P^0_i \}$ 
}
$P^*\gets P^0_i$

\tcp{Evolution loop} 

\For{$t = 0,\ldots, T$}{
    $B^{t+1} \gets \{\}$ \;
    \For{$i = 1,\ldots,M$}{
        $P^t_{k_1}, P^t_{k_2} \gets$ \textrm{sample\_parents}$(B_t)$ \tcp{Sample parents for crossover}

        $P^{t+1}_{i} \gets \mathcal{LLM}(P^t_{k_1}, P^t_{k_2}, s(P^t_{k_1}; \mathcal{D}), s(P^t_{k_2}; \mathcal{D}), p_o, p_c)$         \tcp{crossover operation}

        \If{$u \sim \mathcal{U}(0, 1) < \rho_m$}{
            $P_i^{t+1}  \gets \mathcal{LLM}(P_k^{t+1}, s(P_k^{t+1}; \mathcal{D}), p_o, p_m)$ \tcp{mutation operation}
        } 

        \tcp{critique and simplification}
        $P_i^{t+1} \gets$\textrm{critic}$(P_i^{t+1})$
        $P_i^{t+1} \gets$\textrm{simplifier}$(P_i^{t+1})$

        \If{$s(P_i^{t+1}; \mathcal{D}) > s(P^*; \mathcal{D})$}{
            $P^* \gets P_i^{t+1}$
        }
        $B^{t+1} \leftarrow B^{t+1} \; \cup \; \{ P^{t+1}_i \}$ 
    }
}

\Return $P^*$;

\end{algorithm}

\subsection{Evolutionary Search for Programs}\label{ssec:evolution}
To search through the vast, discrete space of programs, \disciple~adapts evolutionary search.
Evolutionary program search typically starts with a large population of random programs.
These programs are then sampled based on their fitness as parents.
The parent programs create new programs through crossover and mutation, resulting in a new population.
Newer generations improve over the previous as the population is getting optimized for the fitness function.
We use the metric $\mathcal{M}$ as the fitness function in our work.

We keep the overall evolutionary algorithm the same but replace key steps with an LLM.
First, at the start of the process, we provide the LLM with a prompt for the objective to generate the initial programs.
To leverage the prior knowledge of LLMs, we use a prompt $p_o$ that mentions the specified description of the quantity of interest:  ``\emph{Given a satellite image, write a function to estimate $\langle descr \rangle$}''. 
We do not expect the LLM to answer such a difficult scientific question without leveraging the observations $\mathcal{D}$; however, the prompt prevents the evolutionary algorithm from searching in completely random directions. 
As a result, our initial population is not entirely random.

Second, rather than using the symbolic methods of crossover and mutation, we use the LLM to perform these operations.
LLMs have common sense about programming and result in much better program modifications when performing crossover and mutations. 
More specifically, let $P^t_{k_1}$ and $P^t_{k_2}$ be two programs sampled from the $t^{th}$ generation selected as parents based on the fitness function. 
To perform a crossover operation we pass, the objective prompt $p_o$, the two programs $P^t_{k_1}$ and $P^t_{k_2}$, their corresponding scores (using \cref{eq:score}), along with a crossover prompt $p_c$ to obtain a new program:
\begin{equation}
P^{t+1}_{k} = \mathcal{LLM}(P^t_{k_1}, P^t_{k_2}, s(P^t_{k_1}; \mathcal{D}), s(P^t_{k_2}; \mathcal{D}), p_o, p_c)    
\end{equation}

The crossover prompt instructs the LLM to make use of the two-parent program and come up with a new program. 
The LLM is able to combine elements from the parents to produce something new as can be seen in \cref{fig:pipeline}. Please refer to the supplementary 
for more examples.

Similar to crossover, we also mutate a program with some probability using a mutation prompt $p_m$. 

\begin{equation}
^{m}P_k^{t+1} = \mathcal{LLM}(P_k^{t+1}, s(P_k^{t+1}; \mathcal{D}), p_o, p_m)
\end{equation}

We present the exact input fed to the LLM for crossover and mutation in the supplementary. Note that both crossover and mutation operations also include the objective prompt preventing the LLM from generating programs far away from the objective. %

\subsection{Feature Set Prediction}
\label{ssec:setpred}
Solutions for many problems require combination of multiple feature. 
Optimizing for the correct combination of these features is challenging for an LLM, as we are not doing gradient-based optimization.
Therefore instead of prompting the LLM to directly generate a predictor for $y$, we prompt it to create a list of predictive features. 
We then learn a linear regressor on top of this list and use the regressor with the list of features as the final program. 
Since both the generated program and regressor are interpretable, our final program is also interpretable.

\subsection{Program Critic}\label{ssec:critic}
The only form of supervision our method gets is through the metric score $s(P;\mathcal{D})$ created by evaluating the program $P$ on the observations. 
However, since we perform crossover and mutation through a language model, we can provide finer-grained information to LLM in order to aid the search.

More specifically, we propose a critic that performs a finer-grained evaluation of the program that we get after crossover/mutation.
In most visual domains, the data can categorically distributed by using the same set of foundational primitives.
Our critic performs a \emph{stratified evaluation} by partitioning the observation data into multiple categories and evaluating the program on individual strata.
\begin{equation}
\mathcal{D} = d_1 \cup d_2 \cup \cdots \cup d_c, \quad \text{where } d_i \cap d_j = \emptyset \text{ for } i \neq j    
\end{equation}

Since all the problems in our benchmark are geospatial, we use a critic that takes the satellite image corresponding to each input and uses a segmentation model to partition the observation dataset into land-use categories. 
The critic obtains per-partition score $s(P; d_i)$ and prompts the LLM to improve the program on categories the model is bad. The addition of a critic improves the programs on data overlooked by existing programs, resulting in reliable programs.

\subsection{Program Simplification}\label{ssec:simplification}
Successive steps of crossovers and mutations of programs result in large programs with many redundancies, hurting interpretability.
We propose an analytical approach to simplify the programs and remove the redundant parts of it. 
Our generated programs can be represented as a directed acyclic graphs (DAG) (we use the abstract syntax tree (AST); see supplementary).
In these DAGs, all the constants and the arguments of the function are root nodes. 
Only the return statement and the unused variables are the leaf nodes.
Any leaf node that is not a return statement, is a piece of code that is not needed and can be removed. 
We then recursively remove all the leaf nodes that are not return statements.

While removing such nodes (unreachable by the return statement) is useful, there could still be features that are returned by the program but are not contributing. 
Recall that we use linear regression on the list of features returned by the program. The weights assigned to individual features by the regression model are useful indicators of which features are redundant.
We remove the features that have a significantly smaller weight compared to the largest weight in the regression (a threshold of 5\% works well). Removing these features from the return statement results in several newly created leaf nodes. We therefore, redo the recursive leaf node removal to further simplify the program. 
Each generated program is first improved through the critic and then simplified before adding back to the population. Algorithm~\ref{alg:algorithm} shows the complete process.

\section{Results}
\label{sec:results}

\begin{figure}[ht]
    \centering
\includegraphics[width=\linewidth]{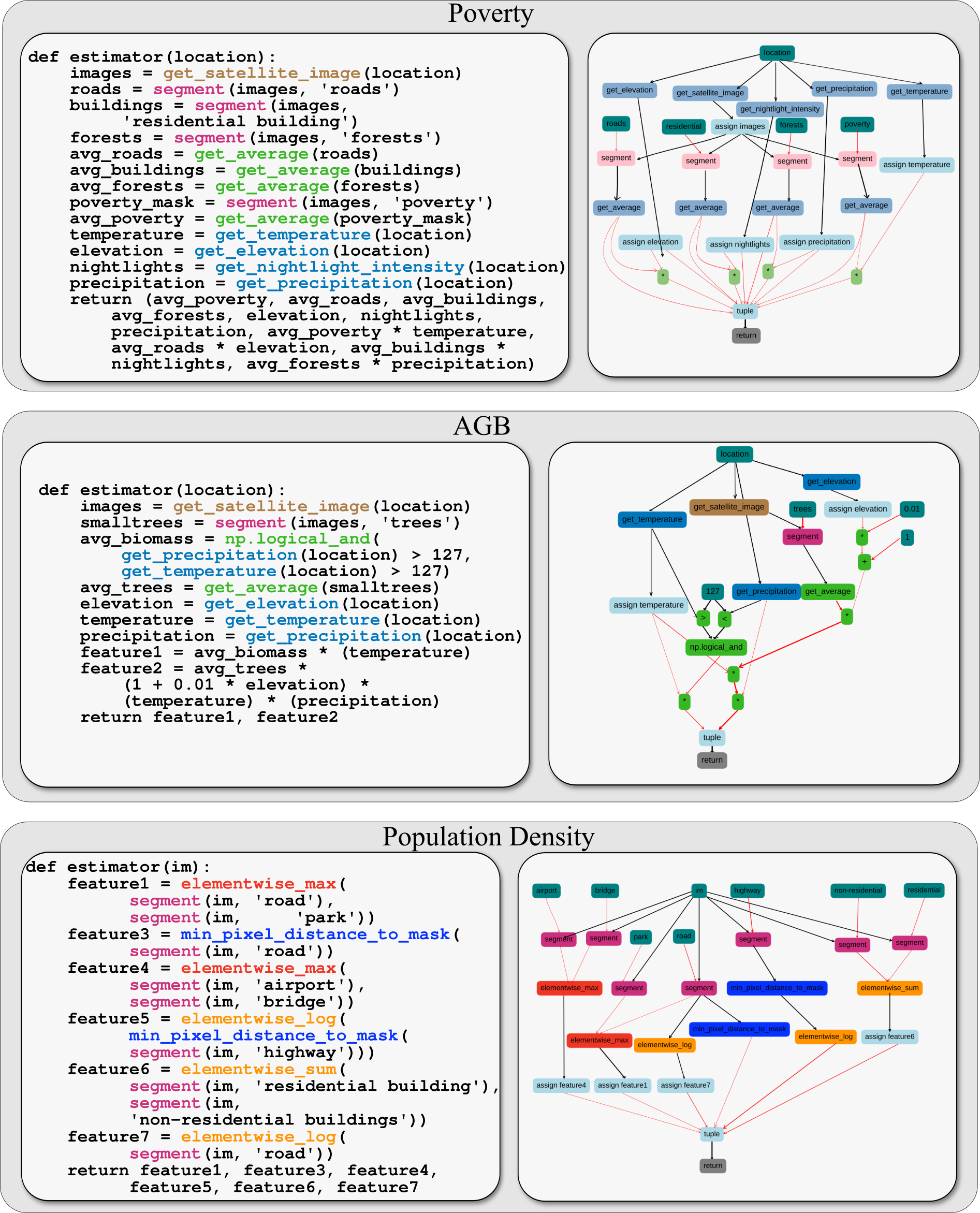}
\caption{The best performing programs for each of the 3 benchmark problems as Python programs (left in each card) and the corresponding DAG representation on the right.
The DAG representation allows better visualization of the importance of different components.
The thickness of the red edges determine how important that component is. 
A black edge represents computation; when removed it is either the same as one of its subsequent edges or removing it could result in a bug.
}
\label{fig:mainresults}

\end{figure}

\subsection{Implementation details}
For all our main experiments we used an open-source LLM \emph{llama-3-8b-instruct}~\citep{dubey2024llama} served using the vLLM library~\citep{kwon2023vllm}.
However, in the supplementary, we also explore other open-source language models.
All the visual data in our benchmarks comes from satellite images, so to allow inferring semantic information from it, we use a black-box open-world foundational model for satellite images, GRAFT~\citep{mall2023graft}. 
Some experiments use ground-truth annotations from OpenStreetMaps~\citep{vargas2020osm} as an alternative to disentangle the effect of segmentation from discovery. 

We run our evolutionary method for $T=15$ generations with a population size of $M=100$. 
For all the problems, the input observation data comes from different geographical locations around the world. 
We split this data into three parts. Two-thirds of the easternmost observations are used to create a training-testing split. The remaining one-third of the data is use to evaluate reliability (out-of-distribution generalization). 
We also release this benchmark for future research in this area. 

\subsection{Benchmark for Visual Program Discovery for Scientific Applications}
\label{ssec:setup}
Given the novelty of the visual program discovery task, there exists no pre-existing benchmark.
We define a new benchmark for this task, drawing on scientifically relevant geospatial problems.
Concretely, we choose two different problems in \emph{Demography}: population density and poverty indicators, 
and to a problem in \emph{Climate Science}: for above ground biomass (AGB) estimation.

It is important to note that for these problems, \emph{true relationships between variables of interest are actually unknown}.
As such, an LLM cannot be expected to produce a good program in a zero-shot manner, because it has never seen these relationships before.
This is in contrast to problems like VQA~\cite{suris-23} where the reasoning required to answer a question is well known and we can simply rely on the LLM's world knowledge.
In the case of scientific discovery, actual data is needed to discover the right reasoning.

In the following, we present the observation datasets, metrics, and overview of primitives. 

\subsubsection{Population Density}

\textbf{Observation Dataset}: 
The problem seeks to predict the population density by observing the satellite images of a region~\citep{metzger-22,metzger-24}. 
We obtain the population density values ($y_i$) for various locations in the USA by using ACS Community Surveys 5-year estimates~\citep{acs2024}.
Input observations ($x_i$) are sentinel-2 satellite images at a resolution of 10m~\citep{drusch2012sentinel}. For this experiment, we also use OpenStreetMaps masks~\citep{vargas2020osm} for 42 different land-use concepts (see supplementary) as part of the input.

\textbf{Metric and Primitives:} 
Population density values are aggregated at the county block group level. The predicted population densities are therefore also aggregated at the county block group level. The metric is the per-block group level average L2 error after applying a log transformation.
Along with the arithmetic, and logical primitives (see supplementary)
, we use open-vocabulary segmentation as a primitive.
The segmentation function returns a binary mask for an input concept.

\subsubsection{Poverty Indicator}
\textbf{Observation Dataset:} For poverty estimation, we use data from SustainBench~\citep{yeh2021sustainbench}. 
The dataset contains coordinate location as input and wealth asset index as output.

\textbf{Metric and Primitives:} We use L2 error for each location as the evaluation metric. To obtain semantic land use information about a location, we first define a \emph{get\_satellite\_image} function, that returns a sentinel-2 satellite image for any location. This can be used in conjunction with the open-world satellite image recognition model to obtain semantic information about the world. 
Other than this we also include as primitives functions that return average annual temperature, precipitation, nightlight intensity, and elevation at the input location. 

\subsubsection{Aboveground Biomass}

\textbf{Observation Dataset:}  
Similar to poverty estimation, the observation variables are an input location and the output AGB estimate. We use NASA's GEDI ~\citep{dubayah2020gedi} to obtain the observation value for three US states. 
We use data from Massachusetts and Maine (North-East) as the train/test set and Washington (NorthWest) as the out-of-distibution set.

\textbf{Metric and Primitives:} We use L2 error as the metric and the same primitives as poverty estimation.

\subsection{Experimental Setup}
For the same set of training data we compare our best generated program with a set of baselines. 
\begin{enumerate}
    \item \textbf{Mean:} A naive baseline that use the mean of the training observation as the prediction.
    \item \textbf{Concept Bottleneck (CB):} Similar to \cite{koh20concept,yang2023language,oikarinen2023label}, we first extract a list of relevant features and train a linear classifier on it. 
    This method is interpretable due to the bottleneck, however it is not very expressive (see supplementary).

    \item \textbf{Deep models:} We use deep models such as ResNets~\cite{he2016resnet} as baseline (see supplementary for details). We use a small and large variant for each.
    
    \item \textbf{Zero-shot:} This baseline tests how good would LLMs be on their own in generating programs solely relying on prior knowledge without any observation. Since the generated programs can vary drastically, we report an average of 5 different zero-shot programs.

    \item \textbf{Random Search:} Instead of evolutionary search, this baseline relies on the stochasticity of LLMs to perform a random search. If ~\disciple~ is better at searching, it should do better than random searching for the same number of calls to an LLM.
\end{enumerate}

\subsection{Results and Discussion}

\begin{table*}
\small
\centering
      \caption{Performance of our programs on in-distribution (left) and out-of-distribution (right) observations across various problems in the proposed benchmark. This shows the reliability of programs produced by \disciple~(\best{red} is best and \sota{blue} is second best).
      } \label{tab:performance}      
      \begin{tabular}{l c c c c c c | c c c c c c} 
        \specialrule{.12em}{.1em}{.1em}       
        & \multicolumn{6}{c}{In distribution}
        & \multicolumn{6}{|c}{OOD}\\
        & \multicolumn{2}{c}{Population Density} & \multicolumn{2}{c}{Poverty} & \multicolumn{2}{c}{AGB}
        & \multicolumn{2}{|c}{Population Density} & \multicolumn{2}{c}{Poverty} & \multicolumn{2}{c}{AGB}
        \\
        & L2-Log & L1-Log & L1 & RMSE & L1 & RMSE
        & L2-Log & L1-Log & L1 & RMSE & L1 & RMSE\\
        \specialrule{.12em}{.1em}{.1em}
        Mean 
        & 0.6696 & 0.6540 & 1.613 & 1.836 & 42.15 & 50.65
        & 0.6734 & 0.6561 & 1.591 & 1.844 & 74.15 & 83.02
        \\
        
        CB 
        & 0.8298 & 0.7279 & 1.229 & \sota{1.476} & 26.33 & 33.49
        & 0.7951 & 0.7112 & 1.257 & 1.504 & 44.19 & 63.52
        \\
        
        Deep - Small  
        & 0.4431 & 0.5006 & 1.238 & 1.637 & 30.72 & 37.03
        & 0.6623 & 0.5967 & \sota{1.284} & \sota{1.654} & \sota{35.27} & \sota{53.06} 
        \\
        
        Deep - Large  
        & \sota{0.3974} & \sota{0.4843} & \sota{1.170} & 1.478 & \best{21.15} & \best{27.86}
        & \sota{0.4460} & \sota{0.5115}  & 1.344 & 1.741 & 35.41 & 70.30  
        \\
        
        Zero-shot 
        & 0.4702 & 0.5371 & 1.525 & 1.754 & 38.80	& 46.41
        & 0.7020 & 0.6412 & 1.510 & 1.773 & 55.11 &	64.32  
        \\

        Random Search 
        & 0.4353 & 0.5118 & 1.277 & 1.679 & 29.40	& 36.70
        & 0.6763 & 0.6298 & 1.418 & 1.840 & 42.32 & 52.53  
        \\
        
        \textbf{Ours} 
        & \best{0.2607} & \best{0.3778} & \best{1.077} & \best{1.314} & \sota{24.79} & \sota{32.99}
        & \best{0.3807} & \best{0.4426} & \best{1.134} & \best{1.420} & \best{31.10} & \best{42.93}
        \\
        \specialrule{.12em}{.1em}{.1em}
      \end{tabular} 
\end{table*}

We first test our programs on unseen \emph{in-domain} observations close to the regions used for training (\cref{tab:performance} (left)).
We observe that ~\disciple~outperforms all interpretable baselines.
It can even outperform a deep model in many cases, specifically on population density estimation, while being significantly more interpretable.
\disciple~ also outperforms zero-shot program inference from LLMs.
As discussed before, this is in line with the fact that \disciple is uncovering new relationships that may not be known to us, and by extension, to the LLM.
The performance of random search while better than zero-shot is significantly worse than ~\disciple. 
This shows that~\disciple~ is able to perform a significantly faster search, by reducing the meaningful search space.
Our evolutionary process effectively leverages data to perform this novel discovery.

\paragraph{Are our programs reliable?} 
If a program is reliable it should be able to generalize to other regions. 
\cref{tab:performance} (right) shows DiSciPLE to these baselines on such an out-of-distribution set.
Here our approach outperforms all baselines \emph{including deep networks}, suggesting that due to its interpretable-by-design representation, our method learns a model that can generalize better and overfit less to the in-distribution training data.

\begin{figure*}[ht]
    \centering
\includegraphics[width=\linewidth]{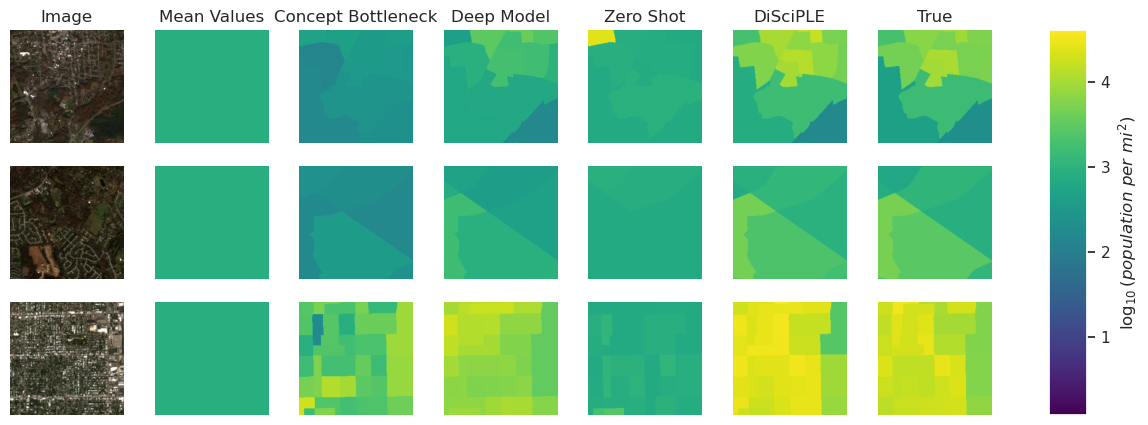}
\caption{Qualitative comparison of ~\disciple~ with other baselines on the tasks of population density. ~\disciple~ Can map to the true population density maps much more accurately than the baselines (Refer to the supplementary for more comparisons). The maps display population density as the base-10 log of people per square mile.
}
\label{fig:qualitative}
\end{figure*}

We also show these results qualitatively in \cref{fig:qualitative}, by comparing population density predictions of ~\disciple~ and the baselines to the true population density.
It is very clearly evident that ~\disciple~ can model the fine-grained changes in population in unseen regions significantly better than the baselines (refer to supplementary for more visualizations).

\begin{figure}[ht]
    \centering
\includegraphics[width=\linewidth]{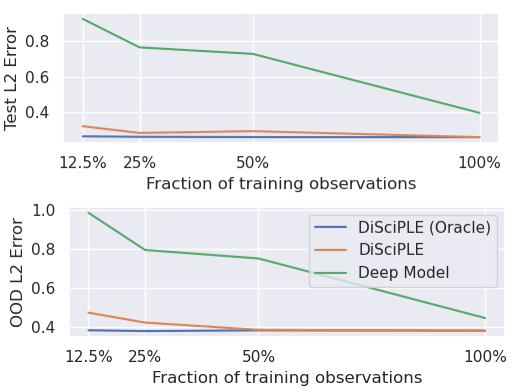}         
\caption{Performance of \disciple~compared to deep baselines as we reduce the amount of training observation (in terms of L2 error). The Oracle (blue) uses a program learned from all observations but uses only partial observation for parameter training. ~\disciple~(orange) uses partial observation during evolution as well. While the errors get worse as we reduce the observation data, the drop is significantly less severe for ~\disciple~ compared to deep models, which tend to overfit.}
    \label{fig:scale}
\end{figure}

\paragraph{Are our programs data-efficient?} 
Our methods are only trained on a maximum of 4000 observations.
\cref{fig:scale} further shows that even when the amount of training data is reduced, our approach shows minimal degradation in performance compared to deep networks.
This suggests that while deep models can learn to generalize with a lot more data, our model does not need as much data to begin with, making it data-efficient.

\paragraph{Are our programs interpretable?} Our programs are interpretable-by-design as we can visualize the factors contributing to performance. 
Fig.~\ref{fig:mainresults} shows such programs (left in each card) for all the problems in our benchmark. 
An expert who is working with our method to figure out such programs can add/edit parts of the formula and figure out which/how much do each of these components matters. 

We perform this step of understanding the influence of individual operations by removing each operation in our program and measuring its effects on the final score. 
The DAGs on the right of each program show the program structure and the red edges show the influence of each component proportional to the width. 
This visualization can allow experts to understand which operations are important for the model.
For example, in the program for population density \cref{fig:mainresults}, we can see that semantic concepts such as ``highway'' and ``residential building'' are very important.

\paragraph{Can our method perform better than expert humans?}
Our method would only be useful in real-world scenarios if it can come up with stronger or comparable programs to human experts.  
We test this on the task of AGB, by providing an expert (a PhD student actively working on AGB) with a user interface with the same information as our method. 
The experts took about 1.5 hours to use their domain knowledge and iterate over their program for AGB estimation. However, the best program they could come up with had an L1 error of \textbf{37.65} on the in-distribution set and \textbf{53.20} on the OOD set (compared to \textbf{24.79} and \textbf{31.10} for ~\disciple). We figure this is primarily because experts need to spend more time on the problem. 
In general, experts would spend numerous days to come up with a good program, 
while our method can come up with a better program faster.

\paragraph{Extension to more indicators}
We also test ~\disciple~ on a larger suite of demographic indicators. Using SocialExplorer, we build a suite of 34 demography indicators. Refer to the supplementary for a list of these indicators. This includes demography information such as age group, education status, etc. In \cref{tab:moreindicators}, we report the average performance of our method compared to baselines on this data.
Since different indicators can have different scales, we first normalize all of them to have zero mean and unit standard deviation. 
These indicators are challenging to predict directly from satellite images, as evidenced by the deep model failing to perform significantly better than CB and mean baselines. 
As a result while ~\disciple~ performs better than all the baselines the improvements are not huge.
Nonetheless, ~\disciple~ performs better than every baseline.
This large-scale experiment shows the potential of applying ~\disciple~ to a wider range of problems. More details about these demographic indicators and individual performance on these is shown in the supplementary.

\begin{table}
\small
\centering
      \caption{Performance of ~\disciple~ compared to baselines on a larger suite of challenging 34 demographic indicators. Since the dataset is very challenging, the deep baseline regresses to mean, however with  ~\disciple~ we can still see some improvements.
      } \label{tab:moreindicators}      
      \begin{tabular}{l c c c c} 
      & \multicolumn{2}{c}{Test} & \multicolumn{2}{c}{OOD} \\
     & L1  & RMSE & L1 & RMSE \\
        \specialrule{.12em}{.1em}{.1em}    
     Mean & 0.8578 & 1.1519 & 0.8939 & 1.1948\\
     CB & 0.8249 & 1.1159 & 0.8771 & 1.1767\\
     Deep & 0.8527 & 1.1556 & 0.8942 & 1.1990\\
     \textbf{Ours} & \textbf{0.8159} & \textbf{1.1065} & \textbf{0.8750} & \textbf{1.1719}\\
        \specialrule{.12em}{.1em}{.1em}       
      \end{tabular} 
\end{table}

\subsection{Ablations}

\paragraph{How important is the role of feature-set prediction, critic, and simplification?}
Table~\ref{tab:ablationparts} measures the performance of our model on the task of population density as we successively add these components to the evolutionary algorithm. 
The addition of feature set prediction instead of a single feature helps, as it allows our method to learn expressive linear regression parameters instead of letting the LLM come up with them.
Further adding critic results in further improvement as the programs start covering nicher concepts resulting in better unseen and OOD generalization. 
Finally adding in simplification also improves the program. 
We posit that simplification removes irrelevant features preventing the LLM from focusing on them when performing crossovers.

\begin{table}
\small
\centering
      \caption{Performance of our method as we successively remove the components. Both critic and simplification lead to performance improvement for our method.
      } \label{tab:ablationparts}      
      \begin{tabular}{c c c c c c c} 
      & & & \multicolumn{2}{c}{Test} & \multicolumn{2}{c}{OOD} \\
      Set & Critic & Simpli. & L2 log & L1 log & L2 log & L1 log \\
        \specialrule{.12em}{.1em}{.1em}       
        \xmark & \xmark & \xmark & 0.3159 & 0.4296 & 0.4835 & 0.5178\\    
        \cmark & \xmark & \xmark & 0.2906 & 0.4049 & 0.4258 & 0.4826\\    
        \cmark & \cmark & \xmark & 0.2873 & 0.3984 & 0.4184 & 0.4684 \\    
        \cmark & \cmark & \cmark & \textbf{0.2607} & \textbf{0.3778} & \textbf{0.3807} & \textbf{0.4426}\\    
        \specialrule{.12em}{.1em}{.1em}       
      \end{tabular} 
\end{table}

\paragraph{How important are common sense and prior knowledge of LLMs?}
The two major advantages an LLM provides over traditional tree-search are: 1) better crossover and mutation as LLMs can understand the meaning of the primitives. 2) use of prior knowledge for better-guided search. 
Therefore we remove these two sources of information and test how well can our method perform. 
To remove the understanding of functions we rename them with meaningless terms and remove the descriptions.
To remove the context of the problem we remove the objective prompt.
\cref{tab:ablationcommonsense}, show the performance of our method on density estimation after removing each of these prompts. 
Without common sense, the search cannot even progress away from the initial random programs, resulting in worse-than-mean results (L1 error of 0.84 vs 0.26 for ~\disciple).
This suggests that symbolic regression models, that have no understanding of open-world primitives, would struggle to search.
If we just remove the context of the problem, the model does slightly better and can obtain results better than the mean and zero-shot programs (L1 error of 0.45). This suggests that while the search is moving in the objective's direction, it is slow.

\begin{table}
\small
\centering
      \caption{Perfomance of our method when removing the context of the problem (objective prompt from the evolution, and when renaming and not describing the primitive functions to the LLM.
      We see significant drops in performance in both cases, suggesting that both common sense and prior knowledge of LLM are important to perform efficient evolutionary search.
      )} \label{tab:ablationcommonsense}      
      \begin{tabular}{c c c} 
      Method & L1 log & L2 log \\
        \specialrule{.12em}{.1em}{.1em}       
        No common-sense & 0.8401 & 0.7186\\    
        No problem context & 0.4498 & 0.5140\\    
        \disciple~full & \textbf{0.2607} & \textbf{0.3778}\\    
        \specialrule{.12em}{.1em}{.1em}       
      \end{tabular} 
\end{table}

\section{Discussion and Conclusion}
\label{sec:conclusion}
\paragraph{Limitations:}One of our limitation is that we can only differentiably optimize learnable parameters in the last computational layer. 
This could miss out on programs with useful parameters in some intermediate computation layers. 
We attempted to make the whole pipeline differentiable, however the model performance did not improve much. 
Many of the operations in our pipeline even though differentiable have zero-gradient in large part of input space, making gradient optimization challenging.
Moreover, a completely differentiable programs is even slower to optimize resulting in much slower evolution.
In future work, we plan to use initialization tricks for non-linear optimization and second-order optimization to obtain even more expressive models.

\paragraph{Conclusion:}We present DiSciPLE -- an evolutionary algorithm that leverages the prior-knowledge and common sense abilities of LLMs to create \emph{interpretable, reliable and data-efficient} programs for real-world scientific visual data. 
This allows us to create programs that are more powerful than existing interpretable counterparts and more insightful than deeper uninterpretable models.
We shows its prowess on 3 scientific applications by proposing a benchmark for visual program discovery.
We believe that using DiSciPLE in tandem with a human expert can rapidly speed up the scientific process and result in numerous novel discoveries.

\paragraph{Acknowledgements}
This research is based upon work supported in part by the Office of the Director of National Intelligence (Intelligence Advanced Research Projects Activity) via 2021-20111000006, the NSF STC for Learning the Earth with Artificial Intelligence and Physics, the U.S. DARPA ECOLE Program No.\ \#HR00112390060. and NSF grants 2403016 and 2403015.
The views and conclusions contained herein are those of the authors and should not be interpreted as necessarily representing the official policies, either expressed or implied, of ODNI, IARPA, DARPA, or the US Government. The US Government is authorized to reproduce and distribute reprints for governmental purposes notwithstanding any copyright annotation therein.

We also thank Pierre Gentine, Juan Nathaniel, Rong-Yu Gu, and Aya Lahlou for their valuable assistance with the expert experiments and for providing the climate science data.
% \clearpage
{
\small
\bibliographystyle{ieeenat_fullname}
\bibliography{indest}
}
\clearpage
\appendix

\end{document}